\documentclass[10pt,twocolumn,letterpaper]{article}

\usepackage[pagenumbers]{cvpr} 

\usepackage{graphicx}
\usepackage{amsmath}
\usepackage{amssymb}
\usepackage{booktabs}
\usepackage{multirow}
\usepackage{adjustbox}

\usepackage[pagebackref,breaklinks,colorlinks]{hyperref}

\usepackage[capitalize]{cleveref}
\crefname{section}{Sec.}{Secs.}
\Crefname{section}{Section}{Sections}
\Crefname{table}{Table}{Tables}
\crefname{table}{Tab.}{Tabs.}

\def\cvprPaperID{40} 
\def\confName{AI4CC}
\def\confYear{2023}

\begin{document}

\title{We never go out of Style: Motion Disentanglement by Subspace Decomposition of Latent Space} 

\author{Rishubh Parihar \textsuperscript{1} \quad Raghav Magazine\textsuperscript{2} \quad Piyush Tiwari\textsuperscript{1} \quad R. Venkatesh Babu\textsuperscript{1} \\
\textsuperscript{1} Vision and AI Lab, IISc Bangalore \quad \textsuperscript{2} IIT Dharwad}



\maketitle

\begin{abstract}
Real-world objects perform complex motions that involve multiple independent motion components. For example, while talking, a person continuously changes their expressions, head, and body pose. In this work, we propose a novel method to decompose motion in videos by using a pretrained image GAN model. We discover disentangled motion subspaces in the latent space of widely used style-based GAN models that are semantically meaningful and control a single explainable motion component. The proposed method uses only a few $(\approx10)$ ground truth video sequences to obtain such subspaces. We extensively evaluate the disentanglement properties of motion subspaces on face and car datasets, quantitatively and qualitatively. Further, we present results for multiple downstream tasks such as motion editing, and selective motion transfer, e.g. transferring only facial expressions without training for it. 
\end{abstract}

\begin{figure}[h]
  \centering
   \includegraphics[width=1.0\linewidth]{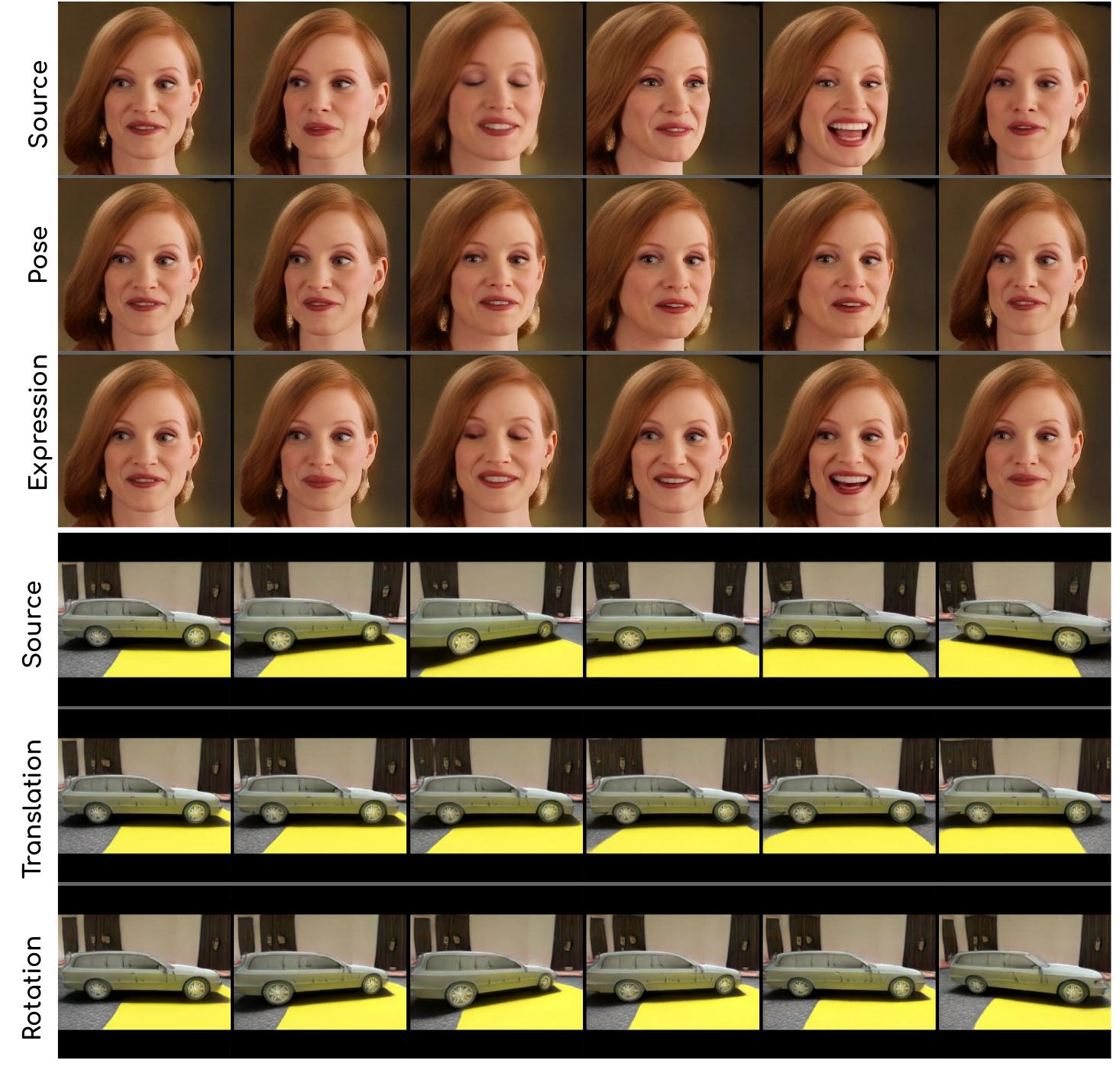} 
   \caption{We perform motion decomposition of a given complex motion into interpretable motion components. Given a source video of a talking mouth, we decompose the motion into disentangled pose and expression motion. Likewise, we decompose the entangled motion of a car into translation and rotation components. Observe the yellow marking on the road moving back to the source and decomposed translation video.} 
   \label{fig:teaser_fig}
   \vspace{-6mm}
\end{figure}

\begin{figure*}[t]
  \centering
   \includegraphics[width=0.8\linewidth]{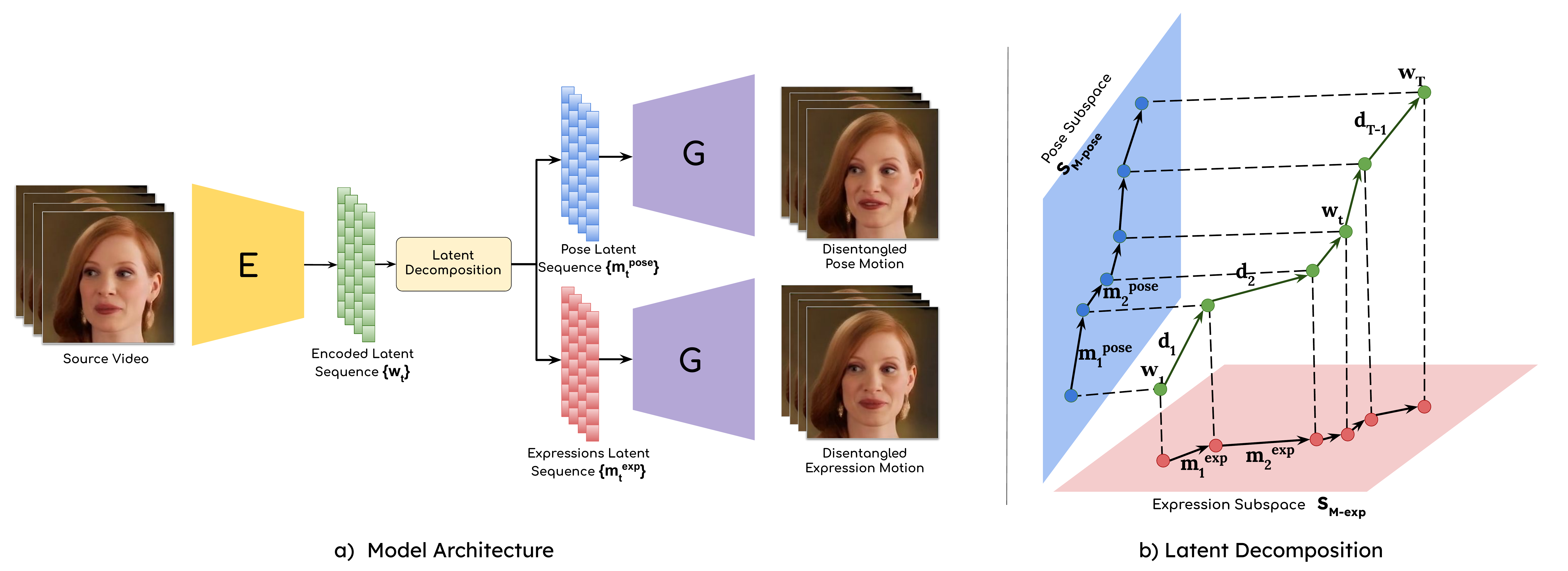} 
   \caption{The motion decomposition methodology consists of three steps - Firstly, the input source video is encoded into a latent sequence using the e4e~\cite{e4e} encoder model. Secondly, the latent trajectory is mapped onto disentangled motion subspaces. Lastly, the projected latent trajectory is processed by the generator to produce disentangled motion videos.}
   \vspace{-4mm}
   \label{fig:method_fig}
\end{figure*}

\section{Introduction}
\label{sec:intro}
Style-based GAN models ~\cite{karras2019style, karras2020analyzing, stylegan-ada, stylegan-v-cvpr2022} have shown unprecedented performance in image generation and control with semantically rich latent spaces. This has fostered research in image editing by latent space manipulation ~\cite{abdal2021styleflow,harkonen2020ganspace,patashnik2021styleclip,shen2020interpreting,parihar2022everything,karmali2022hierarchical}. Furthermore, recent works ~\cite{tzaban2022stitch, latentTransformer, xu2022temporally} extend these capabilities to edit the contents of the video. Specifically, these methods use GAN inversion to first encode the video frames into a sequence of latent codes, which are then edited individually and reconstructed to produce video output. Fundamentally, videos can be decomposed into two major factors: motion and content editing ~\cite{mocogan2018,mocoganhd-iclr2021}, both of which are essential for controlled video generation. While video  content can be edited by extending image editing methods to video frames, motion editing is considerably more complex and necessitates a novel approach.  

In this work, we adopt a new perspective toward motion editing through motion decomposition into interpretable motion components (Fig.~\ref{fig:teaser_fig}). Most real-world motions are intertwined in nature, such as a person's head pose and expressions during speech. Decomposing such complex motion into interpretable components creates novel opportunities for motion editing, such as selectively transferring a single motion component, altering the strength of a particular motion, or removing unwanted motion like camera jitter. 

We use pretrained image GAN ~\cite{karras2019style,karras2020analyzing} models and repurpose them for motion decomposition by leveraging their rich latent spaces. Specifically, we discover low-dimensional latent subspaces in the latent space that control disentangled and explainable motion components. For example, we found subspaces corresponding to pose and expression motion for faces and rotation and translation for cars, as shown in Fig.~\ref{fig:teaser_fig}. By navigating in these subspaces, diverse variations of a single motion can be generated. We learn these subspaces from a dataset of a few disentangled videos with a single motion component with a \textbf{\textit{total duration of $<$15 mins}}. 

Extensive experiments on multiple datasets for motion decomposition prove the effectiveness of the proposed method. Further, we present multiple downstream motion editing tasks - \textit{selective motion transfer, altering the strength of motion components, and stabilizing a jittery video}. Additionally, our method can generate high-resolution outputs and seamlessly integrates with StyleGAN based image manipulation techniques. We summarize the major contributions of our work as follows: 

\vspace{-4mm}
\begin{itemize}
    \item A technique to decompose motion in videos into interpretable components by projecting onto the disentangled motion subspaces in the $\mathcal{W}+$ space. 
    \vspace{-3mm}
    \item A method for fine-grained manipulation of motion in high-resolution videos using pretrained image GANs and only a few minutes of disentangled video.

    \vspace{-3mm}
    \item Multiple downstream applications of motion editing -including motion editing and controlled motion transfer on multiple datasets.
\end{itemize}

\section{Methodology}
\label{sec:method} 
\vspace{-2mm}
\noindent 
The style-based GAN models ~\cite{karras2019style,karras2020analyzing} have semantically rich and disentangled $\mathcal{W/W+}$ latent spaces. Given the smoothness of these latent spaces, a video can be represented as a sequence of latent codes. When this sequence is decoded with the generator, it results in a sequence of corresponding video frames ~\cite{tzaban2022stitch, latentTransformer}. We work on this compressed representation and project the latent trajectory onto interpretable motion subspaces to obtain disentangled motion sequences as illustrated in Fig.~\ref{fig:method_fig}.

\textbf{StyleGAN for Video Representation.} 
Given an input video $X$ consisting of $T$ frames $\{X_t\}$, we encode it into a sequence of latent codes $\{w_t\}$ with a pretrained encoder $\mathcal{E}$ ~\cite{e4e}. The transition vector between consecutive latent codes, $d_t = w_{t+1} - w_t$, represent the motion between the frames $X_t$ and $X_{t+1}$. We process the transition sequence $D = \{d_t\}$ and obtain the edited transition sequence $\tilde{D} = \{\tilde{d_t}\}$. Note that the transition sequence does not have information about the video content and only captures the motion between frames. Finally, we add elements of $\tilde{D}$ to the starting latent code $w_0$, and pass it through a pretrained generator $\mathcal{G}$ to obtain an edited video as a sequence of frames.



\textbf{Motion Decomposition.} 
Our goal is to decompose input transition sequence $D$ representing a complex motion into interpretable transition sequence $D^{M_j}$, where $M_j$ is a given motion component, e.g., pose or expression. We define subspace $S^{M_j} \subset \mathcal{W+}$, which encapsulates the latent vectors corresponding to a single motion component $M_j$. To obtain $D^{M_j}$ with only motion $M_j$, we project the sequence $D$ onto $S^{M_j}$. We formulate the motion subspaces $M_j$s as a linear span of atomic motion vectors (details in the next section). To obtain the decomposed motion video, we added the projected sequence $D^{M_j} = \{d^{M_j}_t\}$ to the starting  latent code $w_0$ and passed it through $\mathcal{G}$.

\textbf{Motion Subspace Estimation.} 
We used a dataset of $N$ ground truth videos having a single motion component $M_j$ (explained in Sec.\ref{sec:impl_det} and supp. document). We accumulate the transition vectors from all the videos in the dataset to obtain a dataset $B_{M_j}={\{{\{\mathbf{d^n_t}}\}_{t=1}^{T}}\}_{n=1}^{N}$ of the transition vectors. $B_{M_j}$ captures all possible motion variations in $M_j$. To approximate the motion space for $M_j$, we perform PCA on the accumulated transition dataset $B_{M_j}$. Specifically, we formulate the subspace $S_{M_j}$ as a linear span of the first $K$ principal components. 

\vspace{-3mm}
\begin{equation}
    S_{M_j} = span(v^{M_j}_0, v^{M_j}_1, v^{M_j}_2, ..., v^{M_j}_K) 
\end{equation}
\vspace{-5mm}

To obtain a projection of transition sequence $D$ onto $S_{M_j}$, we take a dot product between the transition vectors ${d_j}$ and the principal components.

\vspace{-2mm}
\begin{equation}
    d^{M_j}_t = \sum_{i=0}^{i=K} <d_t,v_i> v_i 
\end{equation}
\vspace{-3mm}

Further, we can combine projected trajectories for multiple motions by taking a weighted average of ${d^{M_j}_t}$ with weights $\alpha_j$. Notably, $\alpha_j$ provides fine control over the strength of motion component $M_j$ (see supp. document). We provide ablation of $\alpha_j$ in the video supp. material. 

\vspace{-2mm}
\begin{equation}
\label{eq:combine} 
    \tilde{d_t} = \sum_{M_j} \alpha_j * d_t^{M_j} 
\end{equation} 
\vspace{-4mm} 

\textbf{Motion Transfer.} A natural extension of our framework is controllable motion transfer from a driving video $X^{drv}$ to a source image $I_s$. First, we obtain the edited transition sequence $\{\tilde d\}^{drv}_t$ following Eq.\ref{eq:combine} and add it to source latent code $w_{src}$ to obtain reenacted latent sequence. Our framework provides fine-grained control over the motion transfer and allows us to only transfer a single motion component (ref Fig.\ref{fig:face_motion_decomposition_oneside}) and edit the strength of motion.

\section{Experiments} 
\vspace{-3mm}
\label{sec:expts}
\noindent
To demonstrate the effectiveness of our method, we evaluate the motion decomposition of talking face into expression and pose and moving cars into translation and rotation. Further, we present results on selective motion transfer and ablation studies. Please watch the supp. video to see the video results. 

\vspace{-2mm}
\subsection{Implementation details and datasets}
\vspace{-2mm}
\noindent
\label{sec:impl_det}
We used StyleGAN2 ~\cite{karras2020analyzing} and e4e encoder ~\cite{e4e} models in all our experiments. We used pretrained models for faces and train StyleGAN2 and e4e models for cars on a synthetic cars dataset \textbf{CarsInCity} (ref supp. doc.). To obtain the videos with only pose variations, we use Eg3D ~\cite{eg3D}, to generate random faces with multiple camera positions. We obtain the expression videos from the video expression recognition dataset GRID ~\cite{grid_dataset2006}. For cars, we synthesized the car videos with disentangled rotation and translation in the synthetic blender environment of CarsInCity. Notably, we used videos of total length $<15$ mins to obtain each motion subspace. We used $100$ videos from CelebV-HQ ~\cite{celebv} dataset as a test set which contains high-resolution videos of celebrity talking faces. Additional details are details about the dataset are provided in the supp. document.

\begin{figure}[!t]
  \centering 
   \includegraphics[width=1.0\linewidth]{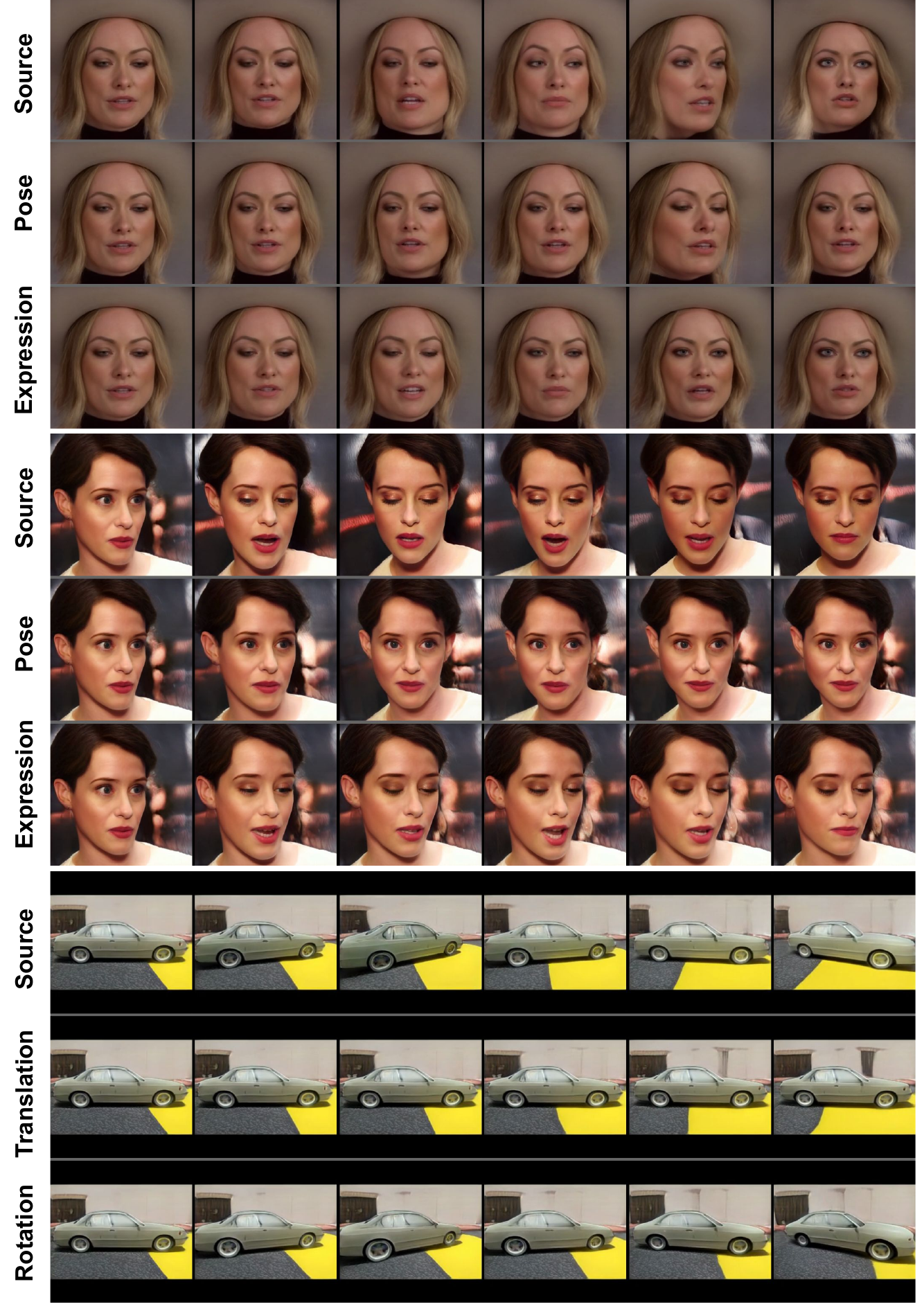}
   \vspace{-7mm}
   \caption{Motion decomposition of face motion into pose and expression and decomposition of car motion into translation and rotation components. Observe the yellow markings on the road moving backward in the source and translation video sequence.} 
   \label{fig:face_motion_decomposition_oneside}
   \vspace{-6mm}
\end{figure} 

\vspace{-2mm}
\subsection{Motion Decomposition}
\vspace{-2mm}
\noindent
We present motion decomposition results generated by our method in Fig.~\ref{fig:face_motion_decomposition_oneside} for faces \& cars. Specifically, observe that the eyes and mouth regions are unchanged throughout the pose subspace, and the pose is intact in all the frames in the expression subspace. Similarly, we can observe that the motion is well disentangled into car rotation and translation components. This suggests that the proposed motion decomposition method generalizes to these complex scenarios. Note that our method has not seen any video of entangled motion during the subspace estimation and is still able to decompose them into disentangled realistic motions. 

\textbf{Qunatitative evaluation.} We quantitatively compare our motion decomposition method with two approaches Independent Component Analysis - \textbf{ICA} and Latent Image Animator - \textbf{LIA} ~\cite{lia_animator}. We performed ICA decomposition of a transition sequence obtained from $100$ videos from CelebV-HQ ~\cite{celebv} with entangled expressions and pose motion. We then annotate each of these directions to either pose or expression and select $6$ components for each motion subspace. For LIA, we analyze the learned motion dictionary of latent codes trained for reenactment. We annotate the vectors in the motion dictionary to either pose or expression based on their alignment with pose or expression motion components. 

We quantitatively evaluate the motion decomposition for faces in two major aspects - \textit{motion disentanglement} and \textit{reconstruction ability}. For motion disentanglement, we defined a novel metric based on the amount of head pose motion: \textit{Aggregated Pose Motion (APM)} which measures the amount of pose change across frames in a given video sequence (ref supp. doc.). Intuitively, when there is only head pose motion, APM should be higher; conversely, if the head pose is fixed and only expressions change, the APM will be smaller. Additionally, we report metrics for Cosine Similarity (CS) on face-recognition embeddings extracted from deep model ~\cite{face-rec-2020}, to measure identity preservation.

To evaluate the reconstruction ability, we project the encoded trajectory onto the two motion subspaces and then combine them back using Eq. ~\ref{eq:combine}. We show quantitative results for metrics evaluating reconstruction quality in Tab.~\ref{tab:reconstruct-face}. Our method can reconstruct the original signal indicated by lower LPIPS and higher SSIM values. Additionally, we achieved the highest CS score, indicating better identity preservation quality. 


\begin{table}[]
\center
\caption{Evaluation of motion disentanglement}
\vspace{-3mm}
\label{tab:decomposition}
\begin{adjustbox}{width=0.7\linewidth}
\begin{tabular}{@{}c|ccc@{}}
\toprule
Motion subspace ($M_j$)                                       & Method & CS $\uparrow$  & APM   \\ \midrule
\multirow{3}{*}{Pose}                    & ICA    & 0.978 & 6.79  \\
                              & LIA~\cite{lia_animator}    & 0.974 & 16.39 \\
                            
                                 & Ours    & \textbf{0.980} & \textbf{12.66} \\ \midrule
\multicolumn{1}{c|}{\multirow{3}{*}{Expressions}} & ICA    & 0.978 & 2.48  \\
\multicolumn{1}{c|}{}                     & LIA~\cite{lia_animator}    & 0.975 & 9.13  \\ 
\multicolumn{1}{c|}{}              & Ours    & \textbf{0.982} & \textbf{5.59}  \\ \bottomrule
\end{tabular}
\end{adjustbox}
\vspace{-2mm}
\end{table}

\begin{table}[]
\center
\caption{Evaluation for reconstruction quality}
\vspace{-3mm}
\label{tab:reconstruct-face} 
\begin{adjustbox}{width=0.6\linewidth}
\begin{tabular}{cccc}
\toprule
      \multicolumn{1}{l}{} Method & LPIPS $\downarrow$ & SSIM $\uparrow$ & CS $\uparrow$  \\ \midrule

ICA &   0.22    &    0.75  &  0.974  \\ 
LIA~\cite{lia_animator}  &  0.23     &    0.73  &  0.975 \\
Ours  &  \textbf{0.18}     &  \textbf{0.80}    & \textbf{0.980} \\ 
\bottomrule
\end{tabular}
\end{adjustbox}
\vspace{-3mm}
\end{table}

\begin{figure}[h]
  \centering
   \includegraphics[width=1.0\linewidth]{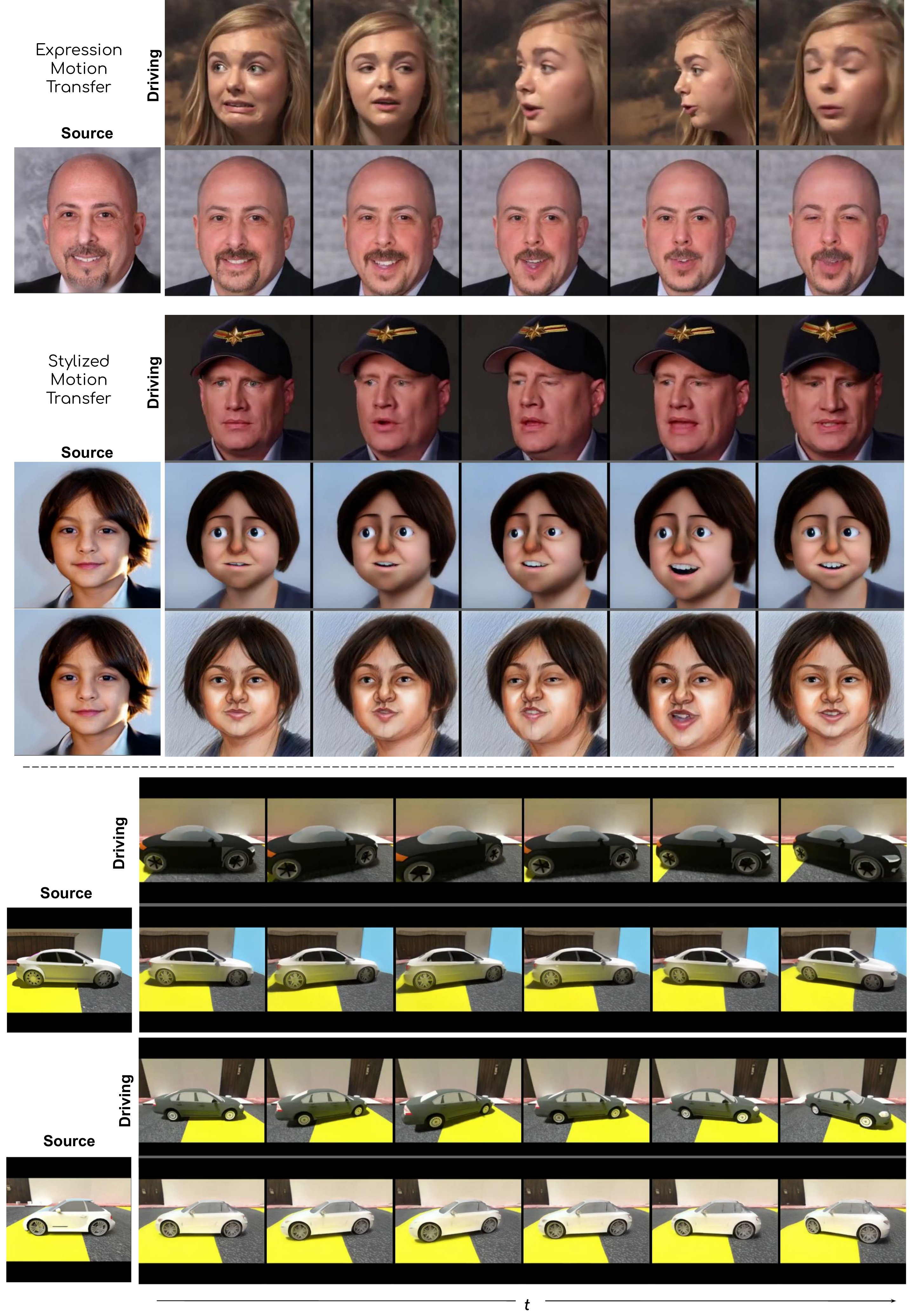} 
   \vspace{-7mm}
   \caption{
   One-shot reenactment results on faces and cars. (Top) Given a driving video of a talking mouth, we transfer only expression and stylized motion transfer on the source image. (Bottom) We transfer the motion of the car rotating on its own axis. Our reenactment is agnostic to the starting pose of the object. Observe that in the last example, although the source car is in the opposite pose, still the rotation motion of the driving video (anticlockwise-clockwise) is transferred correctly.}
   \vspace{-7mm}
   \label{fig:reenactment}
\end{figure} 

\textbf{Ablation on number of Principal Components.}
We ablate over the number of principal components for the pose and expressions subspaces. in Tab.~\ref{tab:n_pc_ablation}. For expression subspace, as we increase the number of components, APM increases, indicating significant pose changes, on the other hand for a small number of components results in a high LPIPS score suggesting a loss of information from the source. For pose subspace, using a larger number of components result in overfitting to unwanted motion that occurred during encoding (ref supp. video). Our experiments found that \#PC=35 for pose and \#PC=50 for expression is a good trade-off between reconstruction and disentanglement.  

\begin{table}[]
\center
\caption{Ablation on the number of principal components}
\vspace{-3mm}
\label{tab:n_pc_ablation}
\begin{adjustbox}{width=0.8\linewidth}
\begin{tabular}{c|ccccc}
\toprule
\#PC-Pose   & 5 & 15 & 35 & 50 & 100 \\ \hline
LPIPS $\downarrow$     & 0.209  & 0.187   & \textbf{0.179}  &  0.174  & 0.158 \\ \midrule
\#PC-Exp    & 5 & 15 & 35 & 50 & 100 \\ \hline
APM &  3.41 &  4.46  &  5.33  & \textbf{5.70}   &  7.00   \\
LPIPS $\downarrow$      & 0.240 & 0.234   & 0.229   & \textbf{0.225}  &  0.213 \\ \bottomrule
\end{tabular}
\end{adjustbox}
\vspace{-6mm}
\end{table}  

\subsection{Motion transfer} 
\vspace{-2mm}
\noindent
We show qualitative results for motion transfer on faces from FFHQ~\cite{karras2020analyzing} and cars from CarsInCity in Fig.~\ref{fig:reenactment}. In the first example, we present results for only expression transfer. In the edited video, expressions are transferred even though the driving frame has extreme head pose motion, demonstrating the high level of disentanglement in the pose and expression subspaces. For the second example in Fig.~\ref{fig:reenactment}, we perform motion transfer and generate the video using StyleGAN-NADA generator for 'Pixar' and 'sketch' domains. Finally, we show motion transfer from rotating cars. The proposed method can perform realistic motion transfer without being trained for the task (ref. supp. video). 

\vspace{-2mm}
\section{Conclusion} 
\vspace{-2mm}
\noindent
This work offers a novel few-shot approach for extracting disentangled motion subspaces within latent spaces. The proposed method enables decomposing of entangled motion patterns in videos into interpretable motion components. Moreover, we demonstrate the versatility of our approach by showcasing its applications in multiple downstream tasks, of motion editing and motion transfer without explicit training. The primary limitation of our work is that it presumes the availability of encoder models of high quality for encoding a video into a latent trajectory.

{\small
\bibliographystyle{ieee_fullname}
\bibliography{egbib}
}

\end{document}